%% file: main.tex
\documentclass[10pt]{article} 
\usepackage[preprint]{tmlr}

\input{math_commands.tex}

\usepackage{hyperref}
\usepackage{url}
\usepackage{graphicx}
\usepackage{amsmath}
\usepackage{amssymb}
\usepackage{booktabs}

\usepackage{multirow}
\usepackage{algorithm}
\usepackage{algpseudocode}

\usepackage{colortbl}
\usepackage{soul}
\usepackage{xcolor}
\usepackage{mdframed}
\usepackage{ulem}
\usepackage{subcaption}
\usepackage{wrapfig}

\title{Sample Weight Estimation Using Meta-Updates \\ for Online Continual Learning }

\author{\name Hamed Hemati \email hamed.hemati@unisg.ch \\
      \addr AIML Lab, School of Computer Science\\
      University of St. Gallen
      \AND
      \name Damian Borth \email damian.borth@unisg.ch \\
      \addr AIML Lab, School of Computer Science\\
      University of St. Gallen}

\def\month{MM}  
\def\year{YYYY} 
\def\openreview{\url{https://openreview.net/forum?id=XXXX}} 

\begin{document}

\maketitle
 
\begin{abstract}
The loss function plays an important role in optimizing the performance of a learning system. A crucial aspect of the loss function is the assignment of sample weights within a mini-batch during loss computation. In the context of continual learning (CL), most existing strategies uniformly treat samples when calculating the loss value, thereby assigning equal weights to each sample. While this approach can be effective in certain standard benchmarks, its optimal effectiveness, particularly in more complex scenarios, remains underexplored. This is particularly pertinent in training ``in the wild,'' such as with self-training, where labeling is automated using a reference model. This paper introduces the Online Meta-learning for Sample Importance (OMSI) strategy that approximates sample weights for a mini-batch in an online CL stream using an inner- and meta-update mechanism. This is done by first estimating sample weight parameters for each sample in the mini-batch, then, updating the model with the adapted sample weights. We evaluate OMSI in two distinct experimental settings. First, we show that OMSI enhances both learning and retained accuracy in a controlled noisy-labeled data stream. Then, we test the strategy in three standard benchmarks and compare it with other popular replay-based strategies. This research aims to foster the ongoing exploration in the area of self-adaptive CL.

\end{abstract}

\section{Introduction}
\label{sec:intro}
Tuning hyperparameters in continual learning (CL) can be crucial for achieving maximum performance in a strategy. Some of the hyperparameters that are often carefully investigated include the learning rate, the mini-batch size, and the coefficient of regularization terms in the objective \cite{Liu2023Online}. For example, the empirical evaluations in \cite{mirzadeh2020understanding} revealed that the choice of hyperparameters such as the mini-batch size and the dropout probability can have a significant impact on the knowledge retention capabilities of a model when trained in a data stream. Furthermore, \cite{han2021convergence} investigated the role of manually designed adaptive learning rates in a CL setting and showed that adaptive learning can mitigate the ``overfitting-to-memory'' issue in replay-based strategies. 

Adaptive optimization of hyperparameters can also be achieved through meta-learning \cite{feurer2014using, razvan2019hyperparameter}. Meta-learning is a learning paradigm in which a learning algorithm improves itself so that it can learn novel tasks more efficiently \cite{dissertation_schmidhuber}. This improvement is made by capturing commonalities between different tasks in a task distribution. For instance, the hyperparameters of an optimization process can be viewed as the meta-parameters of the learning system, which are then updated using feedback from training the model on one or more tasks. Typically, it is assumed that the meta-learner gets full access to a static set of tasks, sampled from a task distribution, in the \textit{meta-training} phase \cite{vinyals2016matching, bansal2021diverse}. However, in the online version of meta-learning, tasks are presented sequentially over time \cite{finn2019online}, and probably not revisited. In general, the goal of online meta-learning is to continuously adapt the meta-parameters in response to the sequence of arriving tasks, enabling it to improve its rapid adaptation to novel tasks. In other words, the adaptation quality of the meta-parameters is expected to improve over time.

Moving along this research path, we also aim to adapt hyperparameters of a learning process based on arriving mini-batches in an online CL stream. In particular, the hyperparameter of interest is \textit{sample importance}. We begin with the assumption that the weight of each sample in the loss function implicitly indicates its ``importance'' in the training process. Intuitively, in the offline setting, the effect of sample importance might appear insignificant, especially given that the model can observe each sample an arbitrary number of times with unrestricted access to the dataset. However, in online CL streams, where the model observes each sample only once, the sample weight can have a substantial impact on the model's learning. Furthermore, in streams where mislabeled samples can appear, for example in self-training \cite{amini2022self}, or self-labeling systems deployed for learning ``in the wild'', a learning system should ideally give lower weight to noisy samples since they could adversely affect the model's performance.   

In theory, all hyperparameters of a learning system can be meta-learned. For example, in \cite{li2017meta}, the authors introduced Meta-SGD to meta-learn optimal learning rates for a particular family of problems. In the context of sample weight estimation, \cite{shu2019meta} developed a method to meta-train a simple MLP that estimates sample weights for the loss function. However, their method is developed for the offline setting and requires a ``meta-dataset'' for applying the meta-updates on the weight estimator network. This implies that, to accurately estimate weight samples in an online CL setting, a form of meta-dataset is required. The meta-objective is then defined as the performance of the adapted model after inner updates on the meta-dataset.

On the other hand, replay-based strategies, which are very popular in CL, utilize a memory buffer to store a small subset of samples from past experiences \cite{de2021continual}. The stored samples are later combined with the arriving data, thus enabling the model to revisit past samples in the face of a distribution shift, and subsequently reduce the forgetting effect on older experiences \cite{chaudhry2019tiny}. In particular, in online CL, the buffer often includes samples from the current experience. 

Our proposed strategy, OMSI, leverages the advantage of the buffer in the experience replay (ER) strategy and uses the buffer as a proxy for the ``distribution observed so far'', to compute the meta-objective. The meta-objective is then used to update the sample importance weights in the mini-batch, which are the meta-parameters of the problem. These updated sample weight estimations enable the model to assign varying weights to each sample during the computation of the loss values in the cross-entropy function, rather than uniformly weighting all samples. The goal is to configure the weights in a way that optimally improves the performance of the current update. 

Below, we summarize our contributions:
\begin{itemize}
    \setlength\itemsep{-0.4em} 
    \item We introduce a novel strategy for estimating the relative importance of individual samples when computing the cross-entropy loss for a mini-batch.
    \item We evaluate our strategy under a setting where labels can be noisy, emphasizing the importance of adaptive sample weight adjustment.
    \item We test the effectiveness of our strategy in enhancing learning outcomes across three standard benchmarks.
\end{itemize}

\section {Related Work}
\label{sec:related_work}

The capacity for adaptation to evolving environments is essential for real-world applications across various domains that necessitate adaptive capabilities  \cite{nagabandi2018learning}. Adaptation can occur in various forms, one of which includes adjusting sample weights in a continual learning process. This section presents an overview of related work in the topics of continual learning and sample weight estimation.

\subsection{Continual Learning with Experience Replay}
Continual learning strategies are gaining increasing importance in image understanding tasks, as demonstrated in \cite{parisi2019continual, de2021continual}. Likewise, natural language processing tasks have also seen significant advantages from CL approaches, as shown in \cite{biesialska2020continual}. In the realm of reinforcement learning, \cite{khetarpal2020towards} have explored the potentials of CL methods in non-stationary environments. Furthermore, in the field of robotics, CL has shown promise, as investigated by \cite{lesort2020continual}.

Replay-based methods in continual learning employ a memory buffer to store samples from past experiences \cite{rolnick2019experience}. The role of replaying previous patterns has also been examined from a neuroscience perspective \cite{mcclelland1995there}. The various aspects of memory-based continual learning have led to a plethora of replay-based strategies \cite{sodhani2022introduction}. For instance, works like \cite{lopez2017gradient, mitchell2018never, de2019episodic} investigate the influence of memory type in CL. Additionally, other research \cite{chaudhry2019tiny, chaudhry2018efficient, riemer2018learning, sprechmann2018memory} focuses on the effective utilization of the memory buffer. Furthermore, some strategies examine the memory population aspects of the replay \cite{aljundi2019gradient, wang2020efficient}. Another research direction in this area involves using generative models to synthesize and replay samples from past experiences alongside data \cite{shin2017continual, sun2019lamol}. Lastly, latent replay, which involves extracting features from a frozen feature extractor and storing them as representations of past experiences for replay, is another approach explored in \cite{ostapenko2022foundational, demosthenous2021continual}.

In the context of scaling sample weights during loss computation for data streams, the most relevant prior work is \cite{guo2020improved}. The authors introduce a method to scale loss for stream and buffer samples using distinct coefficients, determined by a \textit{predefined threshold}. This approach uniformly treats all samples in the buffer and stream, assigning a fixed ratio to each mini-batch chunk. However, it does not explicitly address other facets of negative interference, like that caused by corrupt samples.

\subsection{Continual Learning with Meta-Updates}

Meta-learning methods have been applied in various ways in CL. For instance, Meta-Experience Replay (MER) \cite{riemer2018learning} integrates the concept of meta-replay to both minimize interference and maximize knowledge transfer between consecutive experiences. In the work of \cite{javed2019meta}, the model representations are meta-learned in order to be ``online-aware'' when deployed for inference in online streams. This implies that the model should be able to learn novel classes in the stream without forgetting through the representation space meta-learned in the meta-training phase. In \cite{gupta2020look}, the authors propose LaMAML which employs a look-ahead mechanism through meta-updates to estimate per-parameter learning rates before each update step. By doing this, the learning rate is adapted in a way that minimizes the negative interference between samples. 

Our proposed strategy falls into the same category as LaMAML. These approaches involve applying meta-updates to specific parameters based on the incoming mini-batch of data, followed by executing the final update using the newly updated meta-parameters.

\subsection{Adaptive Learning in Noisy Streams}
Sample weight estimation is commonly investigated under the assumption of noisy labels or unbalanced datasets. For example, \cite{shu2019meta} meta-trains a simple MLP that predicts the weight of a particular sample in the current mini-batch when the dataset is either noisy-labeled or unbalanced in an offline setting. Similarly, another work in this area of research is \cite{ma2018dimensionality} which introduces a noise-tolerant dimensionality reduction technique that remains effective even in the presence of label noise. In the realm of CL, \cite{li2023online} introduces a strategy that leverages both self-supervised and semi-supervised learning to continually learn in a stream that contains mislabeled data. Additionally, another work in this context is based on the idea of ``memory purification'' as explored in \cite{kim2021continual}, which is tested with noisy streams. However, despite the significant advances in this research direction, current methods predominantly rely on heuristic-based approaches for learning from streams with noisy labels.

\section{Preliminaries}
Let $x$ be a random variable taking values in $\mathcal{X} \subseteq \mathbb{R}^d$, which represents the space of all possible inputs with dimension $d$, and let $y$ be a random variable taking values in a finite set of classes $\mathcal{Y}=\{1, 2, \ldots C \}$. Consequently, assume that a classification dataset $\mathcal{D}=\{(x^{(k)}, y^{(k)})\}_{k=1}^{M}$ is provided, where $(x^{(k)}, y^{(k)}) \sim P(x, y)$, with $P(x, y)$ being the joint distribution of $x$ and $y$. Furthermore, the samples in $\mathcal{D}$ are assumed to be \textit{independent and identically distributed}. 

In a supervised learning problem using neural networks, a model $f(x; {\theta})$ is given, where $x$ denotes the input to the model, and $\theta$ is the vector of the model's parameters. The objective is to learn a mapping $f_{\theta}: \mathcal{X} \rightarrow \mathcal{Y}$ that results in minimal empirical risk over the dataset samples. To find the optimal parameters $\theta^* = \underset{\theta}{\text{argmin }} \mathcal{L}(\theta)$, the loss $\mathcal{L}(\theta)$ needs to be computed over the whole dataset. Typically, this loss is computed with equal weights assigned to all samples in the dataset. Since the dataset consists of $M$ samples, the computation of the loss requires a weighted average of the individual sample loss. In the case of uniform sample weighting, the loss is computed as $\mathcal{L}(\theta)=\sum_{i=0}^{M}{ \frac{1}{M} \cdot l(f(x^{(i)}; \theta), y^{(i)})}$,  where $l(.,.)$ is a functional that computes the loss for each prediction.

\subsection{Online Continual Learning}

Moving away from the standard (offline) supervised learning setting, CL introduces a setting where the complete dataset $\mathcal{D}$ is not available at once. Instead, data becomes partially available over time, often in the form of a sequence of experiences $S = [e_1, e_3, ..., e_N]$, referred to as the data stream. In the sequence $S$, each experience $e_i$ contains a train set and a test set, i.e., $e_i=\{\mathcal{D}^{(tr)}_{i}, \mathcal{D}^{(te)}_{i} \}$, where $\mathcal{D}^{(tr)}_{i}$ and $\mathcal{D}^{(te)}_{i}$ are the train set and the test set of the experience respectively. These datasets are assumed to be drawn from the same underlying distributions.

In online CL, each sample within an experience is observed only once. However, computing gradients using only a single sample can be noisy and result in inaccurate gradients in deep models. Therefore, for deep neural networks, a small mini-batch size of $10$ is commonly chosen \cite{mai2022online, Soutif-Cormerais_2023_ICCV,}. This essentially means that the train set of each experience is split into random chunks of mini-batches of size $10$, and each mini-batch is observed exactly once and not revisited. Although the number of epochs in online CL is equal to $1$, the number of \textit{passes} for each mini-batch can be higher, and the model can iterate over each mini-batch multiple times. Additionally, since no ``task indicator'' is used in this setting during training, this online CL setting is often categorized as ``task-agnostic'' \cite{zeno2018task}.

Once the strategy is trained on each experience, we use the test set corresponding to the current experience for evaluation. In this paper, we evaluate the model's performance by calculating the Retained Accuracy (RA) over the test set of all experiences encountered up to the current experience, indicated by $\text{RA}=\frac{1}{N}\sum_{i=1}^{N}\text{Acc}(f_{\theta_{T}^{*}}, \mathcal{D}_{i}^{(te)})$, where $\text{Acc}(f, \mathcal{D})$ represents the accuracy of the function $f$ over dataset $\mathcal{D}$. Another metric that we use is the Learning Accuracy (LA), which is the test accuracy of an experience after training the model on the experience.

\subsection{Hyperparameters as Meta-Parameters}

Meta-learning transforms the optimization problem in the objective into a bi-level optimization problem. The inner optimization problem trains the model according to a given set of meta-parameters and data points from the current \textit{task}. While, the outer optimization problem, updates the meta-parameters after the model is trained through the inner updates. Formally, given a model $f$ with parameters $\theta$, a set of meta-parameters $\lambda$, and a meta-objective function $J(.)$, the goal is to update the meta-parameters after having the model trained on a specific task. In this paper, since the model encounters mini-batches in a data stream, we consider each mini-batch as a ``small task'' for training the model, in order to update the meta-parameters. For clarity, throughout this paper, we will refer exclusively to the data used for updating meta-parameters as mini-batch.

Given that nearly all components relevant to the training process of the strategy can be considered as the meta-parameters of the problem, hyperparameters of the optimization process frequently emerge as a popular choice. Assuming that hyperparameters of the loss function are represented by $\lambda$, the meta-objective is defined as follows:

\begin{center}
    {\textbf{Optimizing Hyperparameters as Meta-Parameters}}
\end{center} 

\begin{equation}
    \begin{gathered}
         \underset{\lambda}{\text{Minimize }}J(f, X_i; \hat{\theta}_i^*, \lambda)  \\
        \text{s.t. } \hat{\theta}_i^* \leftarrow \texttt{InnerUpdates}(f, X_i; \theta_i, \lambda)    
    \end{gathered}
    \label{eq:hypeparameters_metaparameters}
\end{equation}

In Equation \ref{eq:hypeparameters_metaparameters}, the process \texttt{InnerUpdates} performs inner updates using the data from the current mini-batch in order to obtain $\hat{\theta}_i^*$. Once the optimal parameters are derived, the meta-parameters are then adjusted using the feedback from the inner process. An important assumption that we make here is that the entire inner process is differentiable.

Since the meta-parameters are updated using a sequence of mini-batches, this type of problem can be categorized as ``continual learning with meta-updates''. In such a setting, training on each mini-batch is similar to one meta-learning ``session'', where the task data is provided by the mini-batch of the current step. While the primary focus of this paper is on meta-learning the hyperparameters, the choice of meta-objective and other elements in the meta-learning process can vary, depending on the ultimate goal of the learning system.

\section{Method}

In this section, we elaborate on the proposed strategy, OMSI. In OMSI, the meta-parameters are associated with the cross-entropy loss function, a common choice for supervised classification tasks.

\begin{figure}[t]
  \centering
  \includegraphics[width=0.99\textwidth]{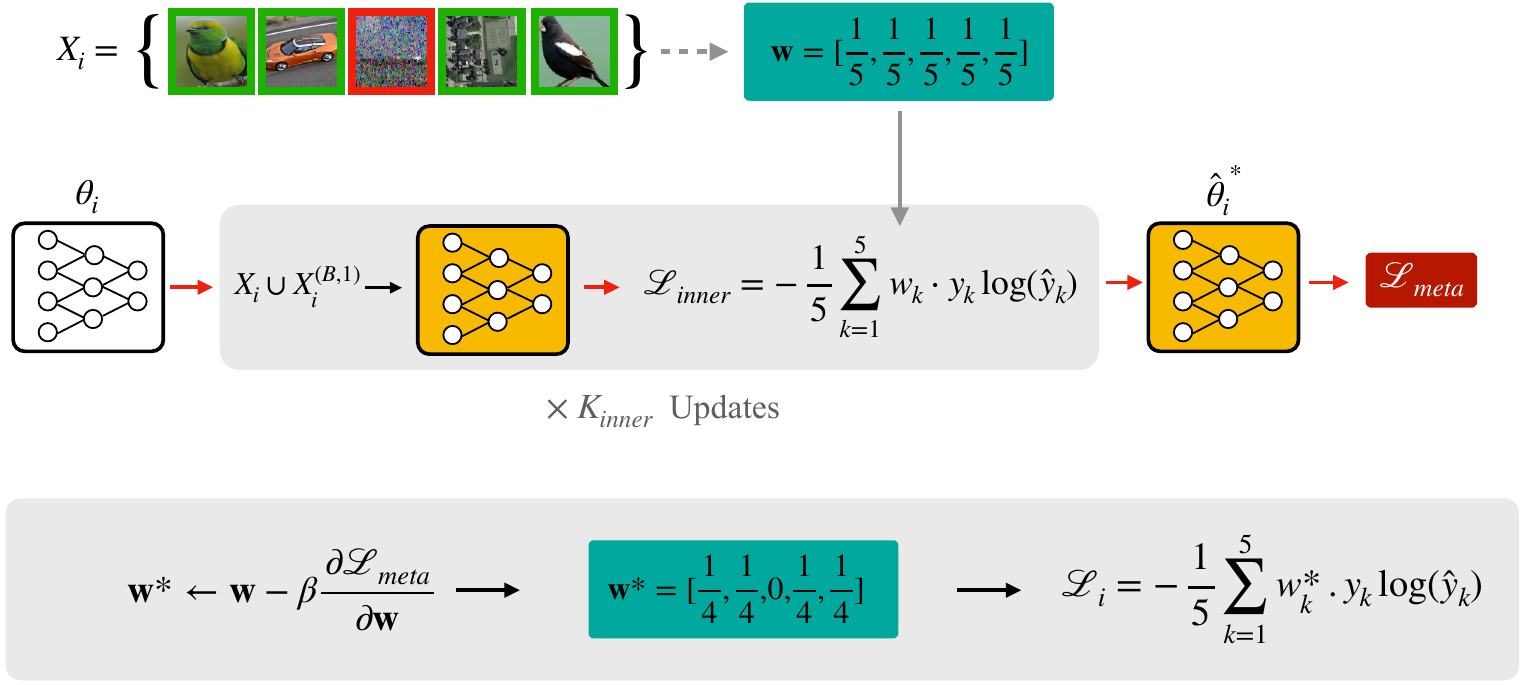}
  \caption[Schematic of the proposed method.]{Schematic of the proposed strategy. (Top) In the first step, the model receives a mini-batch from the stream. After initializing the meta-parameters $\mathbf{w}$, the inner updates are applied to compute the meta-gradients. (Bottom) The sample weights are updated using the meta-gradients to adapt to the current set of samples in the mini-batch.}
  \label{fig:overview}
\end{figure}

The training process of OMSI consists of two primary steps at each iteration $i$. In the first step, we update the meta-parameters associated with the loss function, dynamically adjusting them based on the current mini-batch provided by the stream. Following these adjustments, in the second step, the newly updated meta-parameters serve as tuned hyperparameters for the mini-batch in the optimization stage of the $i$-th iteration. We illustrate the overall process and the schematic of OMSI in Figure \ref{fig:overview}.

In our approach, sample weights are treated as meta-parameters within the learning process, rendering them as learnable variables. Each variable reflects the relative importance of its corresponding sample. Assuming that the current mini-batch at step $i$ consists of $S_i$ samples, denoted as $X_{i}= \{ (x_i^{(1)}, y_i^{(1)}), \ldots, (x_i^{(S_i)}, y_i^{(S_i)}) \}$, we compute the loss value for the $i$-th step as below:

\begin{center}
    {\textbf{Cross-Entropy with Learnable Sample Importance}}
\end{center} 
\begin{equation}
    \mathcal{L}_{ce} = - \sum_{k=1}^{S_i}{\boxed{w_k} \cdot y_i^{(k)} \cdot \log(p(x_i^{(k)}))}
    \label{eq:cross_entropy_learnable}
\end{equation}

In Equation \ref{eq:cross_entropy_learnable}, each $w_k$ represents the weight assigned to each sample within the min-batch. Typically, in the computation of the cross-entropy loss, all samples in the mini-batch are assigned equal weights. As a result, the weight of each sample is equal to $w_k=\frac{1}{S_i}$. 

To apply the first step of OMSI, we begin by initializing the sample weights as a vector. There are two possibilities in terms of how the meta-parameters can be initialized and maintained: (i) to use the updated sample weights from the previous step, or, (ii) to reset the weights at the beginning of each iteration. Given that the value of each variable $w_k$ is ``sample-dependent'', and less important samples can have any index in the mini-batch, it is crucial to reset these values for every mini-batch. Consequently, at each iteration, the vector $\mathbf{w}=[\frac{1}{S_i}, \frac{1}{S_i}, \ldots, \frac{1}{S_i}]$ is initialized as the meta-parameters and is fed to the inner process. Then, an arbitrary number of inner updates are performed, which do not directly modify the original parameters of the model. 

Similar to most replay-based strategies, OMSI combines samples from the current mini-batch with a random subset of samples from the memory buffer. Assuming that $X_i$ is the $i$-th mini-batch from the stream, and $X_i^{(B)}$ refers to the random buffer samples at step $i$, the combined mini-batch is formed as $X_i^{(C)} = X_i \cup X_i^{(B, 1)}$. Since samples from the buffer can be sampled multiple times, $X_i^{(B, 1)}$ indicates the first set of buffer samples that are directly combined with the stream mini-batch. 

Starting from the current model parameters ($\theta_i$), $K_{inner}$ inner updates are applied with the combined mini-batch $X_i^{(C)}$, thereby yielding $\hat{\theta_i}^{(k)}$. The inner steps can be viewed as a ``simulation'' of the training process for the current combined mini-batch, without immediately altering the original model parameters. Once parameters $\hat{\theta_i}^{(k)}$ are obtained, we need to compute the meta-loss. The choice of meta-loss criterion is critical, as it determines the ``upper-level objective''. In OMSI, we set the upper-level objective to find a better sample weight configuration to both reduce negative interference caused by ``corrupt'' samples in the mini-batch during the learning process and ideally improve transfer learning. The key consideration here is that, since there is a probability that the stream mini-batch might contain samples of variable importance or even corrupt samples, we need to compute the meta-loss using ``proxy'' samples. Therefore, we evaluate the model using a combination of $X_i^{(B, 1)}$ with another randomly sampled mini-batch from the memory buffer, denoted as $X_i^{(B, 2)} \sim \mathcal{B}$. 

Following the computation of the meta-loss, we update the meta-parameters $\mathbf{w}$ using the gradients $\frac{\partial \mathcal{L}_{meta}}{\partial \mathbf{w}}$. We apply an SGD update with learning rate $\alpha$ to obtain $\mathbf{w}^*$:

\begin{equation}
    \mathbf{w}^* \leftarrow \mathbf{w} - \alpha \frac{\partial \mathcal{L}_{meta}}{\partial \mathbf{w}}
\end{equation}

Finally, to update the main model, we apply another SGD update on the model's parameters using the adapted sample weights $\mathbf{w}^*$. We present a step-by-step description of the method in Algorithm \ref{alg:omsi_steps},.

\begin{algorithm}[H]
\caption[The algorithm of the proposed method.]{Step-by-step description of the OMSI algorithm for the computation of the cross-entropy loss. The operation $\textbf{SGD-Update}(X, \theta, \mathbf{w})$ performs one step of SGD update on $f$ with samples $X$, parameters $\theta$ and the vector of sample weights $\mathbf{w}$.}
\begin{algorithmic}[1]
\State \textbf{Input:} Model $f$ with parameters $\theta$, update factor $\alpha$, buffer $\mathcal{B}$

\For{$t = 1, 2, \ldots$}
	\State Receive new batch $X_{i}$ from the data stream
    \State \textbf{Initialize:} Model parameters $\hat{\theta}_i^{(0)} \leftarrow \text{Copy}(\theta)$
	\State $X_i^{(B,1)} \sim \mathcal{B}$ \Comment{Random mini-batch from the buffer}
	\State $X_i^{(C)}= X_i \cup X_i^{(B,1)}$ \Comment{Combined mini-batch with $B_i$ samples}
    \State $\mathbf{w} = [\frac{1}{B_i}, \frac{1}{B_i}, \ldots, \frac{1}{B_i}]$
    \For{$j = 1, 2, \ldots, k_{inner}$}
        \State $\hat{\theta_i}^{(j)} \leftarrow$ \textbf{SGD-Update}$(X_i^{(C)}, \hat{\theta_i}^{(j-1)} , \mathbf{w})$
    \EndFor
	\State $X_i^{(B,2)} \sim \mathcal{B}$ \Comment{Random mini-batch from the buffer}
 
	\State $\mathcal{L}_{Meta} = \text{CE}(X_i^{(B,1)} \cup X_i^{(B,2)}; \hat{\theta_i}^{(k_{inner})})$ \Comment{Meta-loss}
	\State $\mathbf{w}^* \leftarrow \mathbf{w} - \alpha \frac{\partial \mathcal{L}_{Meta}}{\partial \mathbf{w}}$ \Comment{Meta-update}
    \State $\theta_i^{(new)} \leftarrow \textbf{SGD-Update}(X_i, \theta_i, \mathbf{w}^*)$
    \State $\textbf{Update-Buffer}(X_i)$
\EndFor
\end{algorithmic}
\label{alg:omsi_steps}
\end{algorithm}

\section{Analysis via Controlled Experiments}

In this section, we conduct three ``controlled'' experiments to assess the effectiveness of adaptive sample importance on learning performance in the presence of artificial label noise, conducted under varying conditions. To generate data streams in the controlled experiments, we employ the online version of the Split-MNIST \cite{swaroop2019improving} benchmark, wherein each experience contains samples from two distinct classes. To make the streams noisy, we add random label noise to the ``even'' mini-batches. The number of noisy samples differs depending on the experiment type. 

It is important to clarify that the primary goal of these experiments is to only examine \textit{the impact of the adaptive learning component} in the learning process, which necessitates the computation of the meta-loss. This computation requires access to ``proxy'' samples from the observed distribution. Therefore, we assume the buffer contains only ``clean'' samples. This essentially means that the buffer needs to be populated with the correct version of the noisy samples in addition to the normal samples. In real-world applications, filling the buffer would require an additional module to verify the quality of each sample before adding it to the buffer. However, in this series of experiments, we are only using the correct samples for updating the buffer without introducing a new component for sample quality check.

Furthermore, given that various factors, such as the weight sample update factor ($\alpha$), the number of inner updates, and the percentage of noisy samples in the noisy mini-batches, can influence the applicability and performance of our strategy, we conduct separate experiments for each aspect. This granular approach allows for isolating the effects of each variable, providing a more detailed understanding of how each factor contributes to the overall performance of the method.

\subsection{Varying the Sample Weight Update Factor}

\begin{figure}[h]
    \centering
    \begin{subfigure}{.48\linewidth}
        \centering
        \includegraphics[width=\linewidth]{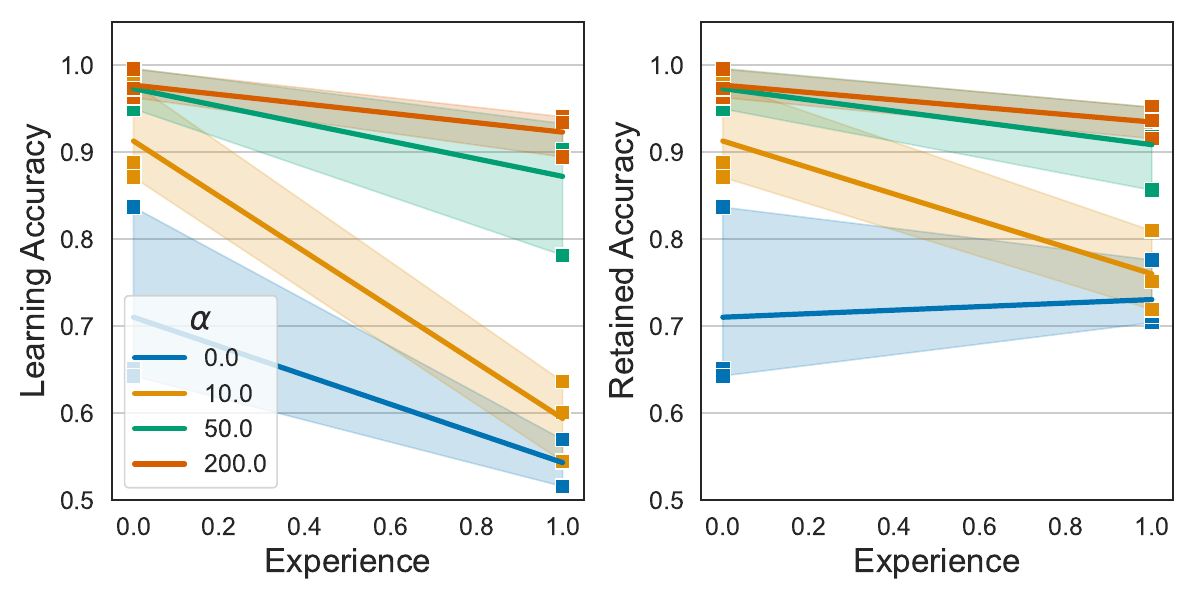}
        \caption{The effect of changing $\alpha$ in a two-experience experiment. By setting $\alpha$ to a value between $[100, 200]$, the strategy effectively adapts the weights of corrupt samples, resulting in better LA and RA values. The results are averaged over three runs with different random seeds.}
        \label{fig:toy_example_alpha}
    \end{subfigure}\hfill
    \begin{subfigure}{.48\linewidth}
        \centering
        \includegraphics[width=\linewidth]{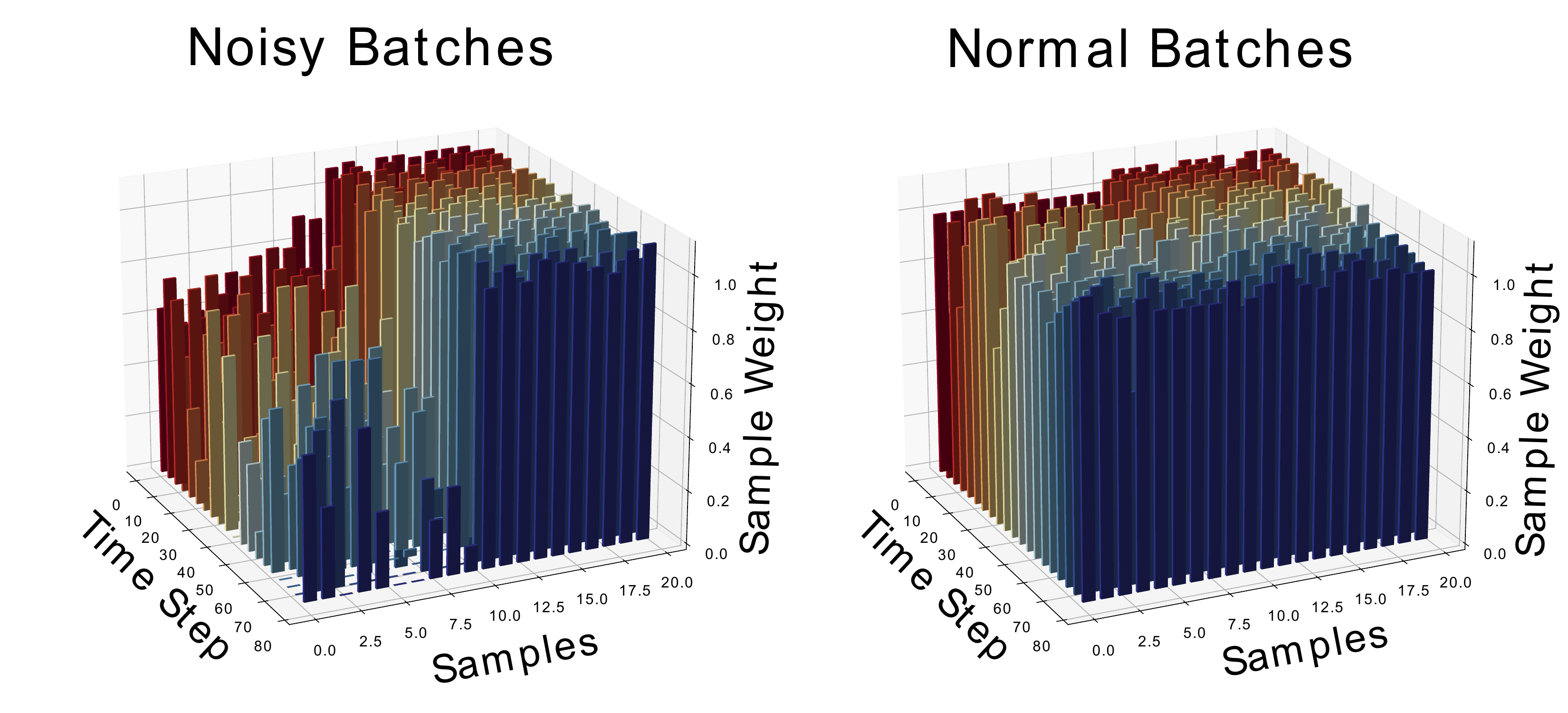}
        \caption{The first $10$ samples are from the data stream, and the remaining $10$ samples are randomly sampled from the buffer. On the left side, noisy mini-batches from even mini-batches are shown. On the right side, normal mini-batches are displayed.}
        \label{fig:toy_example_3d}
    \end{subfigure}
    \caption{Combined figure showing (a) the 3D visualization of normalized sample weights (meta-parameters) and (b) the effect of increasing the alpha factor.}
\end{figure}

In this experiment, the stream consists of only two experiences, and the objective is to understand how different levels of sample weight updates can aid in handling noisy mini-batches during the learning of the second experience. After going through each experience, we evaluate the model's performance in terms of LA and RA. Since the value of the alpha factor $\alpha$ varies, we fix the number of inner updates to $1$, representing the simplest scenario. When $\alpha=0$, the strategy mirrors the ``standard'' Experience Replay (ER) strategy and treats all samples the same way since the sample weights are not changed through the meta-updates. By increasing the $\alpha$ value, we can observe that the effect of adaptive sample weight learning also increases. We show this effect in Figure \ref{fig:toy_example_alpha} where increasing alpha boosts the performance both in terms of LA and RA. Additionally, to investigate whether the noisy-labeled samples in the even mini-batches are assigned with lower weight values, we provide the 3D plot in Figure \ref{fig:toy_example_3d} that demonstrates the sample weights of each sample at each mini-batch. It is evident that the weights of the noisy samples, which are in the even mini-batches, are lower than those of normal samples. 

By giving a lower weight to the corrupt samples, the classification error is reduced and both LA and RA metrics improve. It is important to note that OMSI is task-agnostic, and does not use any information regarding the iteration number and does not use an experience indicator during training. Overall, the results indicate that a higher sample weight update rate can be more helpful, but there are two points to consider here: (i) there is an upper limit to the value that can be used for the update factor, and (ii) this value can be dependent on the dataset.

\subsection{Number of Inner Updates}

\begin{wrapfigure}{r}{0.5\textwidth}
  \centering
  \includegraphics[width=0.48\textwidth]{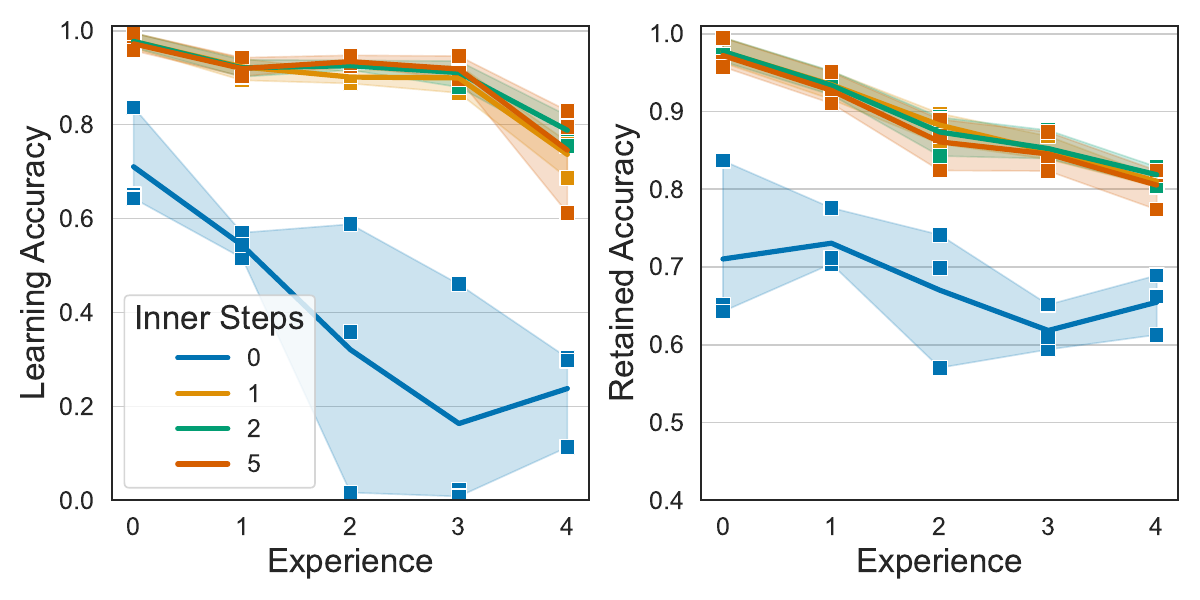}
  \caption{Effect of the number of inner steps on the LA and RA metrics. Increasing the number of inner updates to $2$ or higher does not necessarily result in a significant performance gain.}
  \label{fig:tyo_example_innersteps}
  \vspace{-0.3cm}
\end{wrapfigure}

A determining factor in both the success of OMSI and its computational demand is the number of inner updates. For this experiment, we test whether the number of inner updates over a longer lifetime can have a considerable impact on the final results. For this purpose, we need access to streams with more than two experiences, therefore we use all classes in the modified Split-MNIST benchmark. The streams in this experiment consist of $5$ experiences, where each experience contains data from two classes. Similarly, the noisy-labeled samples appear in the even mini-batches of the stream, and the level of noise in the entire stream is set to $50\%$. This essentially means that all samples in the even mini-batches are assigned with the wrong labels.

Finding the right number of inner updates can be crucial from a computational point of view. A lower number of inner updates means that the method requires less computation and time to obtain accurate meta-gradients, which is often preferred. We can see in the plots demonstrated in Figure \ref{fig:tyo_example_innersteps} that by setting the number of inner updates to $0$, the weight samples do not receive any meta-gradients, effectively rendering the process similar to a standard ER strategy. In that case, the sample weights remain static throughout the learning process, without the possibility for adaptive adjustments based on the incoming data. However, the dynamic nature of our approach becomes evident as we increase the number of inner steps. Furthermore, it is clear that an increasing number of inner updates from from $1$ to $2$ slightly improves the final performance in terms of RA, however, no further improvement was observed by increasing it to $5$ or even higher.

Given that the performance gain from increasing the number of inner steps from $1$ to $2$ or higher is marginal, we opt for a small value of $1$ for inner updates which can still result in a good performance achievement, while minimizing the extra computational demand for the method.

\begin{table}[t]
  \centering
  \begin{tabular}{c c c c }
    \toprule
    \textbf{Fraction} & \textbf{Method} & \textbf{Average Learning Acc. \%} $\uparrow$ &  \textbf{Retained Acc. \%} $\uparrow$ \\
    \midrule
    \multirow{1}{*}{$10\%$} & ER & $\mathbf{94.1 \pm {\scriptstyle 0.6}}$  & $82.0 \pm {\scriptstyle 2.2}$  \\
     & OMSI & $93.8 \pm {\scriptstyle 0.7}$  & $\mathbf{84.4 \pm {\scriptstyle 0.5}}$  \\
    \hline
    \multirow{2}{*}{$20\%$} & ER & $82.3 \pm {\scriptstyle 3.9}$  & $80.1 \pm {\scriptstyle 2.3}$  \\
     & OMSI & $\mathbf{92.4 \pm {\scriptstyle 3.6}}$  & $\mathbf{83.9 \pm {\scriptstyle 1.4}}$  \\
    \hline
    \multirow{2}{*}{$30\%$}  & ER & $76.3 \pm {\scriptstyle 3.6}$  & $73.8 \pm {\scriptstyle 3.8}$  \\
      & OMSI & $\mathbf{94.0 \pm {\scriptstyle 1.0}}$  & $\mathbf{84.2 \pm {\scriptstyle 1.1}}$  \\
    \hline
    \multirow{2}{*}{$40\%$}  & ER & $62.1 \pm {\scriptstyle 6.3}$  & $71.4 \pm {\scriptstyle 2.4}$  \\
      & OMSI & $\mathbf{90.7 \pm {\scriptstyle 2.2}}$  & $\mathbf{78.4 \pm {\scriptstyle 6.4}}$  \\
    \hline
    \multirow{2}{*}{$50\%$}  & ER & $48.2 \pm {\scriptstyle 6.0}$  & $67.8 \pm {\scriptstyle 2.6}$  \\
     & OMSI & $\mathbf{88.6 \pm {\scriptstyle 4.9}}$  & $\mathbf{77.1 \pm {\scriptstyle 8.1}}$  \\
    \bottomrule
    
  \end{tabular}
  \caption{Results for different fractions of noisy samples for the ER and OMSI strategies. The results are averaged over three runs with different class orderings.}
  \vspace{-0.3cm}
  \label{tab:toy_example_noisy}
\end{table}

\subsection{Fraction of Noisy Data}
In this experiment, we aim to investigate how different factions of noisy samples present in each training mini-batch can affect the overall learning process. Therefore, we set the stream mini-batch size and the buffer size to $5$ and $200$, respectively. Similar to the other controlled experiments, we set the ``even'' mini-batches to contain noisy-label samples. Additionally, we use the term \textit{fraction} here to indicate the \textit{ratio} of noisy samples to the total samples in each mini-batch.

We report the results of this experiment in Table \ref{tab:toy_example_noisy}. As we can see in the comparisons, with the percentage of noisy samples in the  mini-batch being on the lower end of the scale, the retained accuracy is not significantly compromised. However, the adaptive weight mechanism continues to provide noticeable benefits for preserving the accuracy levels as we increase the noise level. This suggests that the method exhibits resilience to various levels of noise within the training data in the stream, effectively managing the noise to maintain model performance. These findings hold promise for the practical applications of OMSI in real-world streams as well.

\section{Experiments}

To demonstrate the applicability of the proposed strategy in more practical and widely recognized settings, we conduct several experiments using different benchmarks. For this purpose, we employ three distinct datasets, namely MNIST \cite{deng2012mnist}, CIFAR-100 \cite{krizhevsky2009learning} and Meta Album  \cite{meta-album-2022}, reflecting various domains and characteristics, to create benchmarks for the experiments. The diversity of the benchmarks contributes to the robust testing of OMSI. 

\subsection{Experimental Setting}

\begin{table}[t]
  \centering
  \begin{tabular}{c  c  c c }
    \toprule

    & & \multicolumn{2}{c }{\textbf{Split-MNIST}} \\
    \cline{1-1} \cline{3-4}
    \textbf{Strategy} & & \textbf{Avg. Learning Acc. \%} $\mathbf{\uparrow}$ & \textbf{Retained Acc. \%} $\mathbf{\uparrow}$ \\
    \cline{1-1} \cline{3-4}
    
    Naive & & $\mathbf{94.9 \pm {\scriptstyle 0.3}}$  & $18.6 \pm {\scriptstyle 0.7}$  \\
    ER & & $91.6 \pm {\scriptstyle 2.6}$  & $49.1 \pm {\scriptstyle 5.1}$ \\
    Online EWC \cite{chaudhry2018riemannian} & & $\mathbf{94.9 \pm {\scriptstyle 0.4}}$  & $18.2 \pm {\scriptstyle 1.1}$  \\
    ACE \cite{caccia2021reducing} & & $20.0 \pm {\scriptstyle 0.8}$  & $53.2 \pm {\scriptstyle 2.4}$ \\
    DER \cite{buzzega2020dark} & & $95.1 \pm {\scriptstyle 0.2}$  & $61.4 \pm {\scriptstyle 6.9}$ \\
    MIR \cite{rahaf2019online} & & $71.7 \pm {\scriptstyle 6.7}$  & $30.4 \pm {\scriptstyle 10.1}$ \\
    MER \cite{riemer2018learning} & & $94.1 \pm {\scriptstyle 0.5}$  & $\mathbf{68.8 \pm {\scriptstyle 7.4}}$ \\
    GSS \cite{aljundi2019gradient} & & $93.6 \pm {\scriptstyle 0.1}$  & $42.1 \pm {\scriptstyle 9.1}$ \\
    AGEM \cite{chaudhry2018efficient} & & $94.8 \pm {\scriptstyle 1.0}$  & $23.3 \pm {\scriptstyle 2.2}$ \\
    \cline{1-1} \cline{3-4}
    OMSI & & $91.4 \pm {\scriptstyle 0.8}$  & $63.9 \pm {\scriptstyle 5.7}$ \\
    \cline{1-1} \cline{3-4}
    OMSI (Diff. with ER) & & $\mathbf{\approx - 0.2}$  & $\mathbf{\approx + 14.8}$ \\
    \bottomrule
  \end{tabular}
  \caption[Results for the Split-MNIST benchmark.]{Results for the Split-MNIST benchmark. All memory-based strategies use a buffer size of $50$. Results are averaged over three runs.}
  \label{tab:results_mnist}
\end{table}

For the Meta Album experiments, we use the first $9$ datasets in the ``set-0'' of the meta-dataset to generate CL streams with $9$ experiences, with each experience being associated with one of the datasets. All benchmarks are online and class-incremental, and we fixed the mini-batch size to $10$ for all runs. The number of samples in each class is equal to $40$. In the MNIST and CIFAR-100 streams, the number of classes in each experience is equal to $2$ and $10$ respectively, while in the Meta-Album streams the number of classes in each experience varies depending on the dataset associated with that experience. 

In the MNIST experiments, we use an MLP with one hidden layer as the model architecture, and for both CIFAR-100 and MetaAlbum experiments, we use the standard ResNet-18 model. Since the benchmarks are few-shot, we ran the experiments with the pre-trained version of the model on the ImageNet dataset. To update the model during training, we employed the SGD optimizer, and set the momentum coefficient $0$ since using a non-zero momentum resulted in both lower learning and retained accuracy across all experiments. Moreover, we fix the learning rate to $0.01$ for all experiments. 

To obtain the best hyperparameters for each strategy, we performed a grid search over the possible set of values of each hyperparameter, to ensure that the best values are used. After the grid search, the values that resulted in the highest retained accuracy on the test set of the benchmark were chosen. In OMSI, and all other strategies with episodic memory, we applied the reservoir sampling algorithm for updating the buffer \cite{vitter1985random}. Furthermore, we implemented all experiments using the Avalanche library \cite{JMLR:v24:23-0130}. The source code for our implementation can be accessed publicly at \href{https://github.com/HamedHemati/OMSI}{https://github.com/HamedHemati/OMSI}.

\subsection{Results}

\begin{wrapfigure}{r}{0.5\textwidth}
  \centering
  \includegraphics[width=0.48\textwidth]{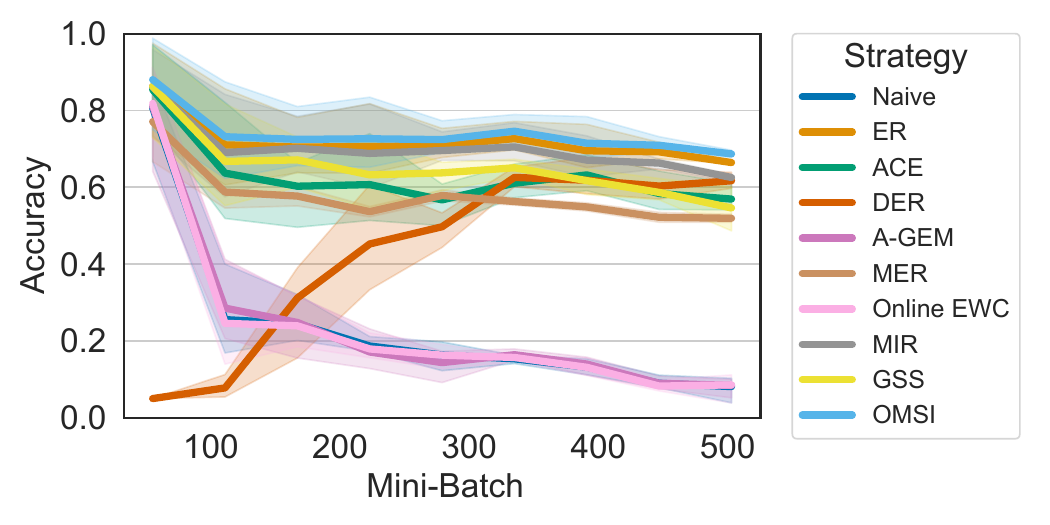}
  \caption[Retained accuracy for all CL strategies in the Meta Album stream.]{Retained accuracy for all strategies in the Meta Album experiment. Each mini-batch is observed only once.}
  \label{fig:results_retained_acc}
  \vspace{-0.5cm}
\end{wrapfigure}

The results from the Split-MNIST benchmark experiments are demonstrated in Table \ref{tab:results_mnist}. We can see that, while MER achieves better performance both in terms of retained accuracy and learning accuracy, it is necessary to highlight that MER is the most computationally expensive strategy among the strategies used for this experiment. It requires a sample-wise meta-update and retrieves buffer samples for every individual sample in the stream mini-batch. In the direct comparison of OMSI with the ER strategy, we observe that the retained accuracy is substantially improved, amounting to $14.8\%$. This improvement, however, only results in a marginal learning accuracy decline. These results indicate that OMSI can be very effective in improving the retained accuracy of ER in the standard Split-MNIST benchmark.

\begin{table*}[t]
  \centering
  \begin{tabular}{c  c  c c }
    \toprule

    & & \multicolumn{2}{c }{\textbf{Split-CIFAR-100}} \\
    \cline{1-1} \cline{3-4}
    \textbf{Strategy} & & \textbf{Avg. Learning Acc. \%} $\mathbf{\uparrow}$ & \textbf{Retained Acc. \%} $\mathbf{\uparrow}$ \\
    \cline{1-1} \cline{3-4}
    Joint & & $-$ & $48.1 \pm {\scriptstyle 0.6}$ \\
    \cline{1-1} \cline{3-4}
    Naive & & $44.6 \pm {\scriptstyle 2.0}$ & $4.9 \pm {\scriptstyle 1.0}$  \\
    ER & & $52.2 \pm {\scriptstyle 1.4}$ & $18.9 \pm {\scriptstyle 1.0}$ \\
    Online EWC & & $45.6 \pm {\scriptstyle 1.3}$ & $4.2 \pm {\scriptstyle 1.1}$ \\
    ACE & & $26.8 \pm {\scriptstyle 1.0}$ & $\mathbf{20.6 \pm {\scriptstyle 0.2}}$ \\
    DER & & $33.6 \pm {\scriptstyle 2.8}$ & $16.2 \pm {\scriptstyle 1.5}$ \\
    MIR & & $52.6 \pm {\scriptstyle 1.0}$ & $17.1 \pm {\scriptstyle 0.5}$ \\
    MER & & $32.2 \pm {\scriptstyle 0.3}$ & $12.0 \pm {\scriptstyle 1.0}$  \\
    GSS & & $52.5 \pm {\scriptstyle 0.3}$ & $18.0 \pm {\scriptstyle 1.0}$ \\
    AGEM & & $45.8 \pm {\scriptstyle 1.6}$ & $4.8 \pm {\scriptstyle 0.5}$ \\
    \cline{1-1} \cline{3-4}
    OMSI (Ours) & & $\mathbf{53.2 \pm {\scriptstyle 1.2}}$ & $19.9 \pm {\scriptstyle 0.8}$  \\
    \cline{1-1} \cline{3-4}
    OMSI (Diff. with ER) & & $\mathbf{\approx +1.0}$  & $\mathbf{\approx + 1.0}$ \\
    \bottomrule
  \end{tabular}
\caption[Results for the Split-CIFAR-100 benchmark.]{Results for the CIFAR-100 experiments. The buffer size is equal to $500$ for all memory-based strategies.}

  \label{tab:results_cifar}

\end{table*}

In the other two benchmarks, CIFAR-100 and Meta-Album, to ensure that the results are not affected by the \textit{combination} of the mini-batches in different strategies in the more, we assigned a dedicated sampler to the data loading process during training. The difference in the combinations can occur due to the difference in the number of update steps in each strategy. This dataloading process guarantees that all models are updated with exactly the same stream mini-batch despite their differences in how the updates are made. 

The CIFAR-100 results, which are shown in Table \ref{tab:results_cifar}, demonstrate that OMSI can also boost performance in the CIFAR-100 streams with approximately $1.0\%$ for both learning and retained accuracy metrics. While the highest retained accuracy is achieved by the ACE strategy, OMSI still achieves a better performance compared to ER and is closer to ACE in terms of RA than ER.

Finally, in the Meta-Album experiments, the OMSI strategy achieves the highest accuracy among all strategies both in terms of learning and retained accuracy. In particular, the retained accuracy outperforms by $1.7\%$. We report these results in Table \ref{tab:results_metaalbum}. Furthermore, to check whether the outperformance with respect to the ER strategy is consistent along the entire stream in the training process, we visualized the retained accuracy of all strategies for the Meta-Album experiments after each experience in Figure \ref{fig:results_retained_acc}. The illustration provides a more granular view of OMSI's performance over time and shows that the outperformance, though marginal, is consistent throughout the entire process.

\begin{table*}[t]
  \centering
  \begin{tabular}{c c c c}
    \toprule

    & & \multicolumn{2}{c}{\textbf{Split-MetaAlbum-V1}} \\
    \cline{1-1} \cline{3-4} 
    \textbf{Strategy} & & \textbf{Avg. Learning Acc. \%} $\mathbf{\uparrow}$ & \textbf{Retained Acc. \%} $\mathbf{\uparrow}$ \\
    \cline{1-1} \cline{3-4} 
    Joint & & $-$ & $84.4 \pm {\scriptstyle 0.5}$ \\
    \cline{1-1} \cline{3-4}
    Naive & & $72.63 \pm {\scriptstyle 0.5}$ & $8.0 \pm {\scriptstyle 3.7}$ \\
    ER & & $76.7 \pm {\scriptstyle 1.0}$ & $66.4 \pm {\scriptstyle 0.5}$ \\
    Online EWC &  & $71.7 \pm {\scriptstyle 0.5}$ & $8.6 \pm {\scriptstyle 4.1}$ \\
    ACE & & $60.7 \pm {\scriptstyle 2.0}$ & $57.0 \pm {\scriptstyle 4.2}$ \\
    DER & & $49.1 \pm {\scriptstyle 4.0}$ & $61.7 \pm {\scriptstyle 2.1}$ \\
    MIR & & $57.6 \pm {\scriptstyle 1.1}$ & $62.6 \pm {\scriptstyle 1.1}$ \\
    MER & & $72.8 \pm {\scriptstyle 0.1}$ & $52.0 \pm {\scriptstyle 1.5}$ \\
    GSS & & $74.9 \pm {\scriptstyle 0.1}$ & $54.7 \pm {\scriptstyle 6.7}$ \\
    AGEM & & $67.6 \pm {\scriptstyle 1.1}$ & $8.4 \pm {\scriptstyle 2.8}$ \\
    \cline{1-1} \cline{3-4}
    OMSI (Ours) & & $\mathbf{77.3 \pm {\scriptstyle 0.8}}$ & $\mathbf{68.1 \pm {\scriptstyle 0.9}}$ \\
    \cline{1-1} \cline{3-4}
    OMSI (Diff. with ER) & & $\mathbf{\approx 0.6}$  & $\mathbf{\approx + 1.7}$ \\
    \bottomrule
  \end{tabular}
\caption{Results for the Split-MetaAlbum experiments. The buffer size is equal to $500$ for all strategies that employ a memory buffer.}

  \label{tab:results_metaalbum}

\end{table*}

Overall, the experimental results in the standard benchmarks also substantiate the feasibility and efficacy of the proposed method in the more commonly used data streams as well. However, the amount of outperformance can vary depending on the underlying dataset.

\section{Conclusion and Future Work}

In this paper, we introduced a novel continual learning strategy, OMSI, that uses the concept of sample importance by modifying the sample weights in the loss function. By employing a buffer that represents a proxy for the observed data distribution at each training step, OMSI computes a meta-loss on each mini-batch of data in the stream and measures how each sample can cause interference with previously seen experiences, without human-designed heuristics. Through a series of experiments on both controlled and standard benchmarks, we demonstrated that the proposed strategy can estimate sample importance in terms of interference with old experiences, and can improve the standard ER strategy.

While OMSI shows promise, it does come with its limitations. In particular, it approximates sample importance using a limited buffer, which may become less accurate over time if the buffer size is small. Additionally, the meta-update can result in extra computational costs. Although the experiments show the efficacy of the approach under certain conditions, it would be advantageous to further test it on larger, more complex real-world datasets, and in different application domains.

Despite the drawbacks, a deeper exploration of various meta-objectives and optimized unrolling techniques in the inner loops could improve the efficiency and accuracy of the method. By pursuing this research avenue, the hope is to continue to advance the state of continual and make progress towards more robust, adaptive learning systems that learn in an ever-changing world without continuous human supervision for tuning hyperparameters. 

\newpage

\bibliography{main}
\bibliographystyle{tmlr}


\end{document}

%% file: math_commands.tex
\usepackage{amsmath,amsfonts,bm}

\def\eqref#1{equation~\ref{#1}}

\def\1{\bm{1}}

\DeclareMathAlphabet{\mathsfit}{\encodingdefault}{\sfdefault}{m}{sl}
\SetMathAlphabet{\mathsfit}{bold}{\encodingdefault}{\sfdefault}{bx}{n}

